\documentclass[sigconf]{acmart}

\AtBeginDocument{%
  \providecommand\BibTeX{{%
    \normalfont B\kern-0.5em{\scshape i\kern-0.25em b}\kern-0.8em\TeX}}}

\setcopyright{acmcopyright}
\copyrightyear{2023}
\acmYear{2023}
\acmDOI{https://doi.org/10.1145/3583780.3615208}

\usepackage{algorithm}
\usepackage{algpseudocode}
\usepackage{algpascal}
\usepackage{bm}
\usepackage{multirow}
\usepackage{xcolor}
\usepackage{balance}

\newcommand\Tau{\mathcal{T}}

\copyrightyear{2023} 
\acmYear{2023} 
\setcopyright{acmlicensed}\acmConference[CIKM '23]{Proceedings of the 32nd ACM International Conference on Information and Knowledge Management}{October 21--25, 2023}{Birmingham, United Kingdom}
\acmBooktitle{Proceedings of the 32nd ACM International Conference on Information and Knowledge Management (CIKM '23), October 21--25, 2023, Birmingham, United Kingdom}
\acmPrice{15.00}
\acmDOI{10.1145/3583780.3615208}
\acmISBN{979-8-4007-0124-5/23/10}

\begin{document}

\title{G-Meta: Distributed Meta Learning in GPU Clusters for Large-Scale Recommender Systems}

\author{Youshao Xiao}
\affiliation{%
  \institution{Ant Group}\country{Hangzhou, China}}
   
\email{youshao.xys@antgroup.com}

\author{Shangchun Zhao}
\affiliation{%
  \institution{Ant Group}\country{Hangzhou, China}}
\email{shangchun.zsc@antgroup.com}

\author{Zhenglei Zhou}
\affiliation{%
  \institution{Ant Group}\country{Hangzhou, China}}
\email{zhouzhenglei.zzl@antgroup.com}

\author{Zhaoxin Huan}
\affiliation{%
  \institution{Ant Group}\country{Hangzhou, China}}
\email{zhaoxin.hzx@antgroup.com}

\author{Lin Ju}
\affiliation{%
  \institution{Ant Group}\country{Hangzhou, China}}
\email{julin.jl@antgroup.com}

\author{Xiaolu Zhang}
\affiliation{%
  \institution{Ant Group}\country{Hangzhou, China}}
\email{yueyin.zxl@antgroup.com}

\author{Lin Wang}
\affiliation{%
  \institution{Ant Group}\country{Hangzhou, China}}
\email{fred.wl@antgroup.com}

\author{Jun Zhou}
\affiliation{%
  \institution{Ant Group}\country{Hangzhou, China}}
\email{jun.zhoujun@antfin.com}

\renewcommand{\shortauthors}{Youshao Xiao et al.}

\begin{abstract}
 Recently, a new paradigm, meta learning, has been widely applied to Deep Learning Recommendation Models (DLRM) and significantly improves statistical performance, especially in cold-start scenarios. However, the existing systems are not tailored for meta learning based DLRM models and have critical problems regarding efficiency in distributed training in the GPU cluster. It is because the conventional deep learning pipeline is not optimized for two task-specific datasets and two update loops in meta learning. This paper provides a high-performance framework for large-scale training for Optimization-based Meta DLRM models over the \textbf{G}PU cluster, namely \textbf{G}-Meta. Firstly, G-Meta utilizes both data parallelism and model parallelism with careful orchestration regarding computation and communication efficiency, to enable high-speed distributed training. Secondly, it proposes a Meta-IO pipeline for efficient data ingestion to alleviate the I/O bottleneck. Various experimental results show that G-Meta achieves notable training speed without loss of statistical performance. Since early 2022, G-Meta has been deployed in Alipay's core advertising and recommender system, shrinking the continuous delivery of models by four times. It also obtains 6.48\% improvement in Conversion Rate (CVR) and 1.06\% increase in CPM (Cost Per Mille) in Alipay's homepage display advertising, with the benefit of larger training samples and tasks.
 
\end{abstract}

\ccsdesc[500]{Computing methodologies~Machine learning}
\ccsdesc[500]{Computing methodologies~Distributed computing methodologies}

\keywords{Recommender System; Deep Meta Learning; Distributed Training}

\maketitle

\section{Introduction} 

Deep Learning Recommendation Models (DLRM) are ubiquitously adopted by internet companies in core applications, such as Advertising, Search, and Recommendations (ASR) scenarios, which influence revenues at billions of dollar level~\cite{cheng2016wide,ma2020temporal,zhou2019deep}. However, learning-based methods are data-demanding and work poorly on users, items, or scenarios with little logging data, which is known as the cold-start problem~\cite{volkovs2017dropoutnet,pan2019warm}. It not only significantly hinders revenue but also largely downgrades the satisfaction of new users or advertisers. 

Meta learning, especially optimization-based meta learning, has recently been widely applied to the cold-start problem in DLRM, effectively alleviating the problem~\cite{du2019sequential, pan2019warm,lu2020meta,huan2022industrial,vartak2017meta}.  The paradigm of optimization-based meta learning~\cite{hospedales2021meta} usually consists of two update loops and datasets: the inner loop is to learn the task-specific parameters using the support set and another outer loop is to update the meta parameters based on the formerly computed parameters using the query set. Among these methods, Model Agnostic Meta Learning (MAML) is the most classical approach and different model derivatives optimize the algorithm varying the base neural network~\cite{ravi2017optimization,bertinetto2018meta}, or the optimization method~\cite{nichol2018first,rajeswaran2019meta}. Additionally, parallelization is essential to train massive datasets with up to billions of samples or model parameters to ensure statistical performance and in-time delivery in the industry~\cite{abadi2016tensorflow,paszke2019pytorch,chen2015mxnet}. 

\begin{figure*}[!ht]%
    \centering
\includegraphics[width=0.9\linewidth]{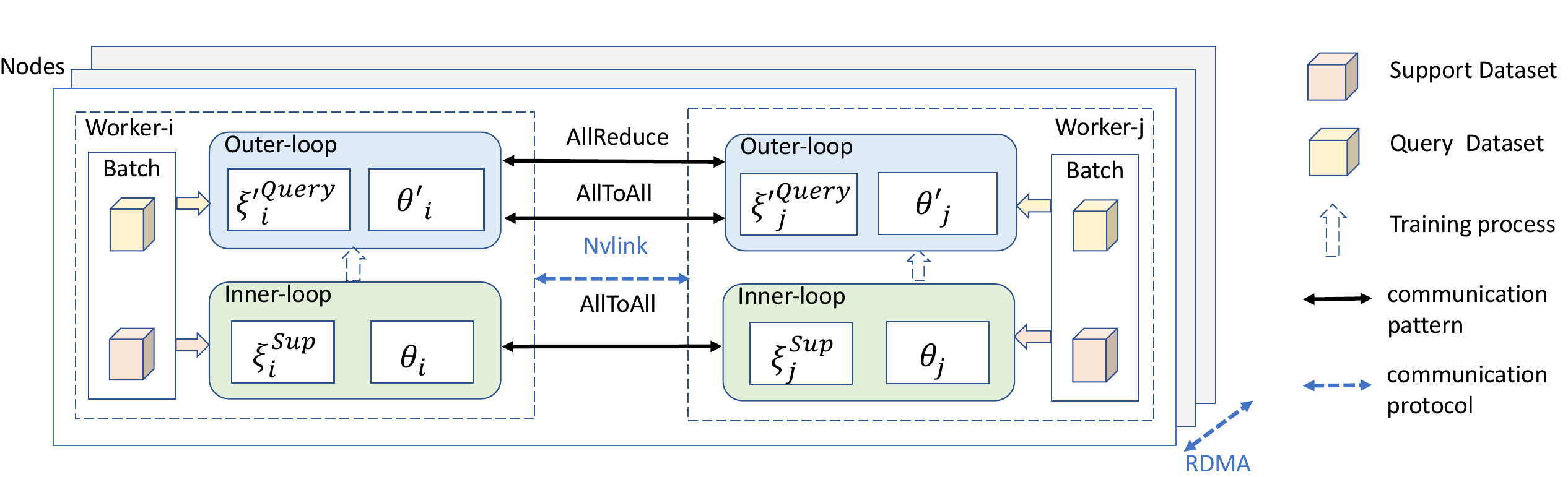} %
    \caption{The Distributed Training Architecture of G-Meta}%
    \label{fig:arch}%
\end{figure*}

However, there are two primary problems when parallelizing the Meta DLRM. The first problem is that the existing distributed training architectures do not fit the recommender system workloads for meta learning paradigms. Parameter Server~\cite{dean2012large,li2014scaling} is one of the most popular architectures for the distributed DLRM training and DMAML~\cite{bollenbacher2020investigating} customizes the Parameter Server architecture for MAML training in the CPU cluster. However, two update loops in meta learning double the computing and the computation-intensive dense layer becomes more complicated in DLRM~\cite{zhang2022picasso,wang2022merlin}, which makes the CPU time-consuming to compute and requires GPU for acceleration. Nevertheless, Parameter Server is mainly used in the CPU cluster and the design underutilizes the capability of GPU since the embedding layers held in servers are I/O and communication-intensive operators~\cite{sergeev2018horovod}. AllReduce~\cite{allreduce,sergeev2018horovod} is another popular architecture in the GPU cluster, and QMAML~\cite{kunde2021distributed} parallelizing the MAML algorithm based on AllReduce, while the GPU memory of a single device cannot hold the huge embedding layer in DLRM\cite{wang2022merlin}. 


Secondly, meta learning requires different data management against the traditional deep learning training, and conventional I/O design bottlenecks the training speed~\cite{aizman2019high,mudigere2022software}. Meta learning requires assembling the batch data according to both the task level and batch level when traditional deep learning only requires batch level in the training pipeline~\cite{abadi2016tensorflow,wang2022merlin,zhang2022picasso}. To be more specific, each worker may hold the batch data from different tasks, but the samples in a batch should belong to the identical task after shuffling for correctness. QMAML~\cite{kunde2021distributed} customizes the data loading in CV scenarios, nevertheless, it is unproductive to ingest massive datasets in ASR scenarios, e.g., billions of KB-level small samples~\cite{aizman2019high}.

Therefore, we propose G-Meta to speed up the distributed training of Optimization-based Meta DLRM in GPU clusters. This work makes three primary contributions: (1) \textbf{Hybrid Parallelism.} G-Meta parallelizes the training of industrial Meta DLRM in GPU clusters by firstly introducing Hybrid Parallelism via \textit{AlltoAll} and \textit{AllReduce} communication primitives. (2) \textbf{Optimization for Parallelization.} G-Meta significantly improves the computation and communication efficiency in distributed meta learning from both algorithm and engineering aspects. (3) \textbf{High-Performance Meta I/O.} G-Meta offers a high-throughput data loading for meta samples by designing the Meta-IO pipeline with I/O optimizations.

\section{Our System}
The G-Meta consists of two components: Meta-Train and Meta-IO. Firstly,  Meta-Train is responsible for high-performance distributed training in the GPU cluster. Secondly, a Meta-IO pipeline is designed to maximize the speed of data ingestion. Without loss of generality, we take the MAML algorithm as an example and it could be easily extended to other optimization-based algorithms. Following notations in MAML paper~\cite{finn2017model}, we present G-Meta in detail.

\subsection{Meta-Train}
The goal of Meta-Train is to learn meta model parameters: embedding layer parameters $\xi^\ast$ and dense layer parameters $\theta^\ast$, where $\xi^\ast$  is too huge to be held in a single device in GPU clusters. 
Firstly, G-Meta evenly partitions the enormous embedding parameters and distributes them to all workers; it shares small dense parameters among all workers. Then it employs a hybrid parallelism algorithm to enable efficient training with careful orchestration. Also, we introduce additional network optimizations from the hardware aspect. The training architecture of G-Meta is depicted in Figure \ref{fig:arch}.

\subsubsection{Distributed Inner Loop}
 Following Algorithm 1, we first parallelize the inner loop by distributing the learning processes over all workers to accelerate the training and save the scarce GPU memory. In the inner loop, MAML only utilizes the mini-batch support set $\mathcal D_{i}^{Sup}$ for model computation and  will utilize the mini-batch query set $\mathcal D_{i}^{Query}$ in the following outer loop. In this case, the Meta DLRM model requires fetching the embedding parameters using the I/O and communication-intensive embedding lookup operations twice, one for the support set and another for the query set, which is inefficient. Therefore, to lower the communication frequency, we aggregate these two embedding lookups and prefetch both embedding parameters  $\xi_{i}^{Sup}$ for the support set and $\xi_{i}^{Query}$ for the query set altogether. During the phase, to fully utilize the bandwidth among workers, we perform the \textit{AlltoAll} primitive~\cite{jeaugey2017nccl} to exchange the embedding parameters among all workers in contrast to parameter server architecture. After that, each worker locally computes the loss function $\mathcal L^{Sup}_{i}$ in the inner forward and updates model parameters in the backward propagation. As the result, each worker $i$ owns its task-specific model parameters ${\xi'_{i}}^{Sup}$ and $\theta'_{i}$.

\subsubsection{Distributed Outer Loop}

In the outer loop, we locally update the embedding parameters from the query dataset $\mathcal D_{i}^{Query}$ which overlaps with the support dataset $\mathcal D_{i}^{Sup}$ using formerly computed ${\xi'_{i}}^{Sup}$ to obtain ${\xi'_{i}}^{Query}$; otherwise, the outer-loop utilizes the stale embedding parameters due to the above prefetch optimization. Then we take the outer forward propagation based on the query dataset and obtain the loss $\mathcal L^{Query}_{i}$ for each worker. We update the huge embedding parameters using the \textit{AlltoAll} primitive and update the small DNN parameters via the \textit{AllReduce} primitive globally. Lastly, we achieve the meta parameters $[\xi, \theta]$ for the next iteration, and repeat the procedure until coverage.

\alglanguage{pseudocode}
\begin{algorithm}[!ht]
\caption{Hybrid Parallelism Algorithm for G-Meta}
\label{algo:maml}
\textbf{Input}:$\alpha$, $\beta$ are step size hyperparameters, $N$ is workers number
\begin{algorithmic}[1]
\Statex \textbf{Worker $i$}:
\State{Randomly initialize $\xi_{i}$ and
$\theta_{i}$ for each worker $i$} \Comment{$\xi$ is bucketized in shards by rows and evenly distributed to workers as $\xi_{i}$. $\theta_{i}$ is the replica of $\theta$ in the worker $i$. }

\While{not coverage}
\State{Sample a batch of task $\Tau_{i} \sim p(\Tau) $ for worker $i$ } 
\State{Split $\Tau_{i}$ into mini-batch support set $\mathcal D_{i}^{Sup}$ and mini-batch query set $\mathcal D_{i}^{Query}$.}
\State{\textbf{Perform \textit{AlltoAll} primitive to prefetch $\xi_{i}^{Sup}$ for  support dataset $\mathcal D_{i}^{Sup}$ and $\xi_{i}^{Query}$ for query dataset $\mathcal D_{i}^{Query}$.}}

\State{Inner forward $\mathcal L^{Sup}_{i} \gets \mathcal L (f(\xi_{i}^{Sup}, \theta_{i};\mathcal D_{i}^{Sup}))$}

\State{Local Backpropagation ${\xi'_{i}}^{Sup} \gets \xi_{i}^{Sup} - \alpha \nabla_{\xi_{i}^{Sup}} \mathcal L^{Sup}_{i}$}

\State{Local Backpropagation $\theta'_{i} \gets \theta_{i} - \alpha \nabla_{\theta_{i}} \mathcal L^{Sup}_{i}$}

\State{\textbf{Update overlapping embedding parameters using  ${\xi'_{i}}^{Sup}$ to get  ${\xi'_{i}}^{Query}$ for outer loop.} }

\State{Outer forward $\mathcal L_{i}^{Query} \gets \mathcal L (f({\xi'_{i}}^{Query}, \theta'_{i};\mathcal {D_{i}}^{Query}))$}

\State{\textbf{Global Backpropagation and communicate via \textit{AlltoAll} primitive: $\xi \gets \xi - \beta  \sum_{i=1}^{N} \nabla_{\xi} \mathcal L^{Query}_{i}$ }}

\State{\textbf{Global Backpropagation and communicate via \textit{AllReduce} primitive ${\theta} \gets \theta - \beta \sum_{i=1}^{N} \nabla_{\theta} \mathcal L^{Query}_{i}$ }}
\EndWhile
\end{algorithmic}

\end{algorithm}

\subsubsection{Optimization for Outer Update Rule}
The outer loop update rule in MAML naturally corresponds to $\theta \gets \theta - \beta \nabla_{\theta} \sum_{i=1}^{N} \mathcal L^{Query}_{i}$ in the parallel setting\cite{kunde2021distributed}. However, it requires a central node to \textit{Gather} all $N$ task-specific parameters from the inner loop and update the meta parameters, which becomes a bottleneck. The data transferred is $K(N-1)$ ($K$ is the data transmission for each node), and the computational complexity is $O(KN)$ in the central node. Nevertheless, we could exchange the order of the gradient and summation operator into: ${\theta} \gets \theta - \beta \sum_{i=1}^{N} \nabla_{\theta} \mathcal L^{Query}_{i}$, according to the derivative rules. Then each worker locally calculates the gradients and aggregates them into meta parameters via the \textit{AllReduce} primitive (e.g., \textit{Ring-AllReduce} ). Thus, the data transferred downsizes to $\frac{2K(N-1)}{N}$ and the computational complexity reduces to $O(K)$. 

\subsubsection{Optimization for Network}
The above hybrid parallelism algorithm enables parallelizing the large Meta DLRM, however, the \textit{AlltoAll} and \textit{AllReduce} primitives are highly-connected communication patterns while the socket-based network in the data center impedes the communication efficiency. Therefore, we further utilize the dedicated RDMA over Converged Ethernet(RoCE) based network in the inter-node data transmission to enable scalable high-speed communication. For the intra-node data transmission, we leverage the NVLink for higher bandwidth instead of the PCIe bus (e.g. system memory). The usage of RDMA and NVLink promotes high-performance data transfers for the highly-connected communications in our G-Meta training.

\subsection{Meta-IO}

In contrast to CPU devices, using GPUs could largely shorten the model computation duration, but it requires high-throughput I/O and network to swallow data faster, otherwise, the training is still bottlenecked. After optimizing the computing and communication phase, we further design and optimize the data ingestion for meta data in the data preprocessing and training phase. Furthermore, the pipeline fits the data ingestion of all Meta DRLM models with high-performance I/O, not just optimization-based meta learning.

\begin{figure}[!ht]%
    \centering
\includegraphics[width=0.9 \linewidth]{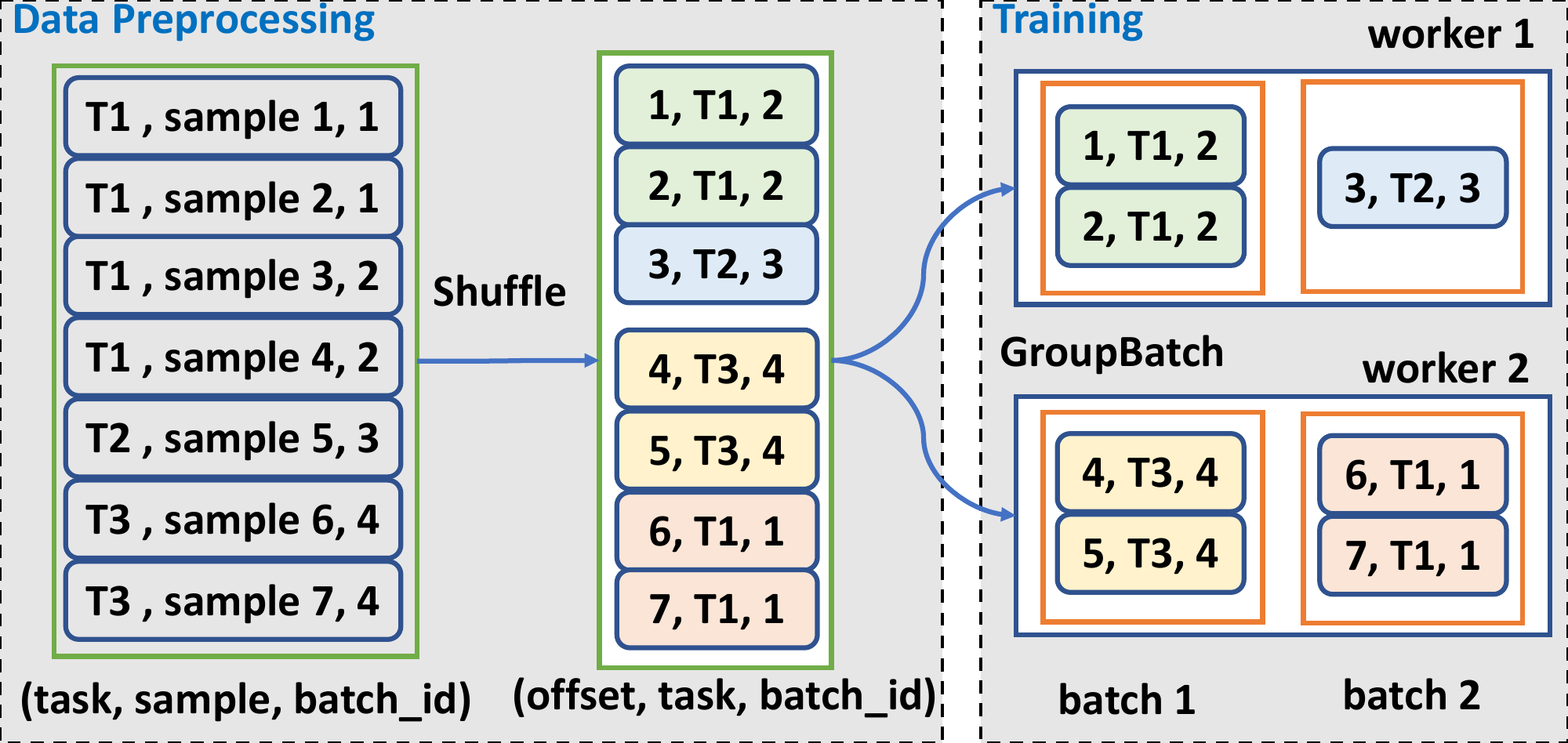} %
    \caption{Dataflow of Meta-IO}%
    \label{fig:meta-io}%
\end{figure}

\subsubsection{Meta I/O pipeline}
Different from conventional deep learning pipeline, the meta data for training requires that the batch of records $\Tau_{i}$ sampled from $p(\Tau)$ should belong to one identical task. To achieve efficient data loading, we first sort the samples by the order of $task$ column, which is the unique id for different tasks, and generate a $batch\_id$ for each sample according to the batch size and $task$ column in the data preprocessing phase as shown in Figure \ref{fig:meta-io}. Secondly, we randomly shuffle the samples with respect to the batch level; otherwise, the traditional sample-level shuffle will mix samples from different tasks and brings unnecessary complexity to data organization. Lastly, in the training phase, samples are loaded by each worker evenly and only records from the same tasks are ensembled in a batch using our \textit{GroupBatchOp} according to both $task$ id and $batch\_id$. We implement the preprocessing phase in MapReduce~\cite{dean2008mapreduce} and the \textit{GroupBatchOp} in the training phase using C++.

\subsubsection{Optimization for Data Ingestion} We further optimize the phase to feed more samples to avoid I/O bottleneck. Due to the massive capacity of samples in ASR scenarios, we have to store samples in the HDD-based file system, such as HDFS\cite{borthakur2007hadoop}, rather than the expensive SSD. To efficiently load massive KB-level small samples, we first generate an extra $offset$ column in the preprocessing phase and sequentially store samples according to the $offset$ column. Then samples could be loaded sequentially in the training phase according to $(offset*i, offset*i+total\_samples/N)$ for each worker $i$. The above sequential read access allows high-throughput I/O in the block-based file system. Secondly, the decoding is time-consuming in the mainstream string-based storage format from our profiling. It incurs considerable latency in the data loading since GPUs with the aforementioned optimizations largely shorten the model computation. Therefore, we utilize the TFRecords\cite{tfrecords} in Tensorflow and WebDataset~\cite{aizman2019high} in PyTorch as the storage format to speed up the unserialization and reduce the data transmission.

\section{Experiments}
\subsection{Experiments Settings}
\subsubsection{Environments Setup}
We conduct experiments over a dedicated CPU cluster and a GPU cluster since using the PS-based method in GPU clusters is ineffective as previously stated. The CPU cluster consists of at most 200 nodes (160 workers and 40 servers), where each worker owns 18 cores and each server owns 22 cores. In contrast, the GPU cluster contains up to 32 Nvidia A100 GPUs. These nodes are connected via RDMA or Socket-based Networks. Similarly, GPUs in each node are connected via NVLink or system memory for comparison. We implement the G-Meta in Tensorflow and employ NVIDIA's collective communications library (NCCL)~\cite{jeaugey2017nccl} for \textit{AlltoAll} and \textit{AllReduce} operations. 

\subsubsection{Experiments Setup}
Three datasets and four models are used for evaluation from two aspects. For statistical correctness, three popular meta DLRM (MAML~\cite{vuorio2019multimodal}, MeLU~\cite{lee2019melu}, and CBML~\cite{song2021cbml}) using Movielens dataset~\cite{harper2015movielens} are leveraged for verifying the correctness of the implementation, following the model settings in TSAML~\cite{yang2022task}. For efficiency, we evaluate G-Meta using an in-house Meta DLRM model on both Ali-CCP dataset~\cite{ma2018entire} and the in-house dataset with 1.6 billion records, since the MovieLens dataset lacks enough samples for large-scale training. For the baseline, we use the DMAML~\cite{bollenbacher2020investigating}, which employs the parameter server architecture for the distributed meta learning training and we also use optimized Meta-IO to avoid I/O bottlenecks for fairness. We run all experiments three times and take average values for evaluation.

\subsection{Experiment Results}
This subsection evaluates the efficiency (training speed or scalability) and statistical performance (AUC) of G-Meta against the DMAML. For the training speed, Table ~\ref{tab:maml} illustrates that G-Meta over $2\times 4$ GPUs is even 22\% faster than DMAML over 160 workers and 40 servers using a total of 3760 cores in the public dataset, which process 169k and 138k samples per second respectively. In this case, G-Meta even achieves cost savings of 62.29\% compared to DMAML, as per the pricing information provided by Aliyun~\cite{aliyun}. Furthermore, G-Meta can process 618k samples per second if we scale out to $8\times 4$ GPUs. Regarding the scalability, the speedup ratio of the DMAML falls while G-Meta's gradually shrinks. The speedup ratio of DMAML is 0.88 and 0.59 when we scale out to 40 and 160 workers. In contrast, the speedup ratio of the G-Meta slowly reduces from 0.94 to 0.86 when we scale out to $2\times4$, and $8\times4$ GPU workers. In the more complicated in-house dataset, we see a similar trend where the speedup ratio using G-Meta declines to 0.88, but the PS-based one drops to 0.58. Lastly, Figure \ref{fig:auc} verifies that G-Meta does not lose the statistical performance.

\subsection{Ablation Study}
\begin{figure}
\begin{minipage}[t]{0.44\linewidth}
\includegraphics[width=\linewidth]{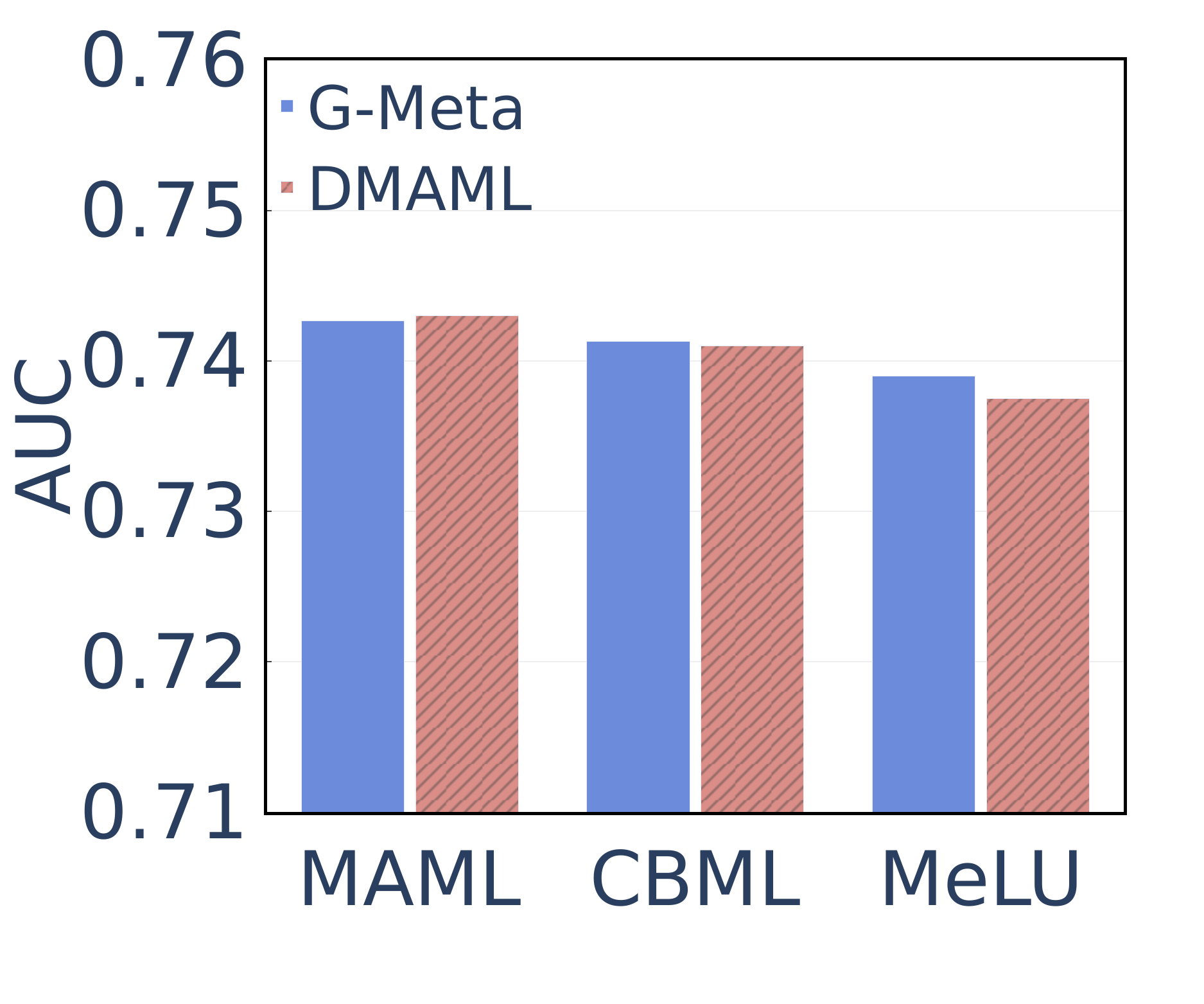}
\caption{Model Performance between G-Meta vs. DMAML using MAML~\cite{vuorio2019multimodal}, MeLU~\cite{lee2019melu}, and CBML~\cite{song2021cbml} in Movielens dataset.}
\label{fig:auc}
\end{minipage}
\hfill
\begin{minipage}[t]{0.45\linewidth}
\includegraphics[width=\linewidth]{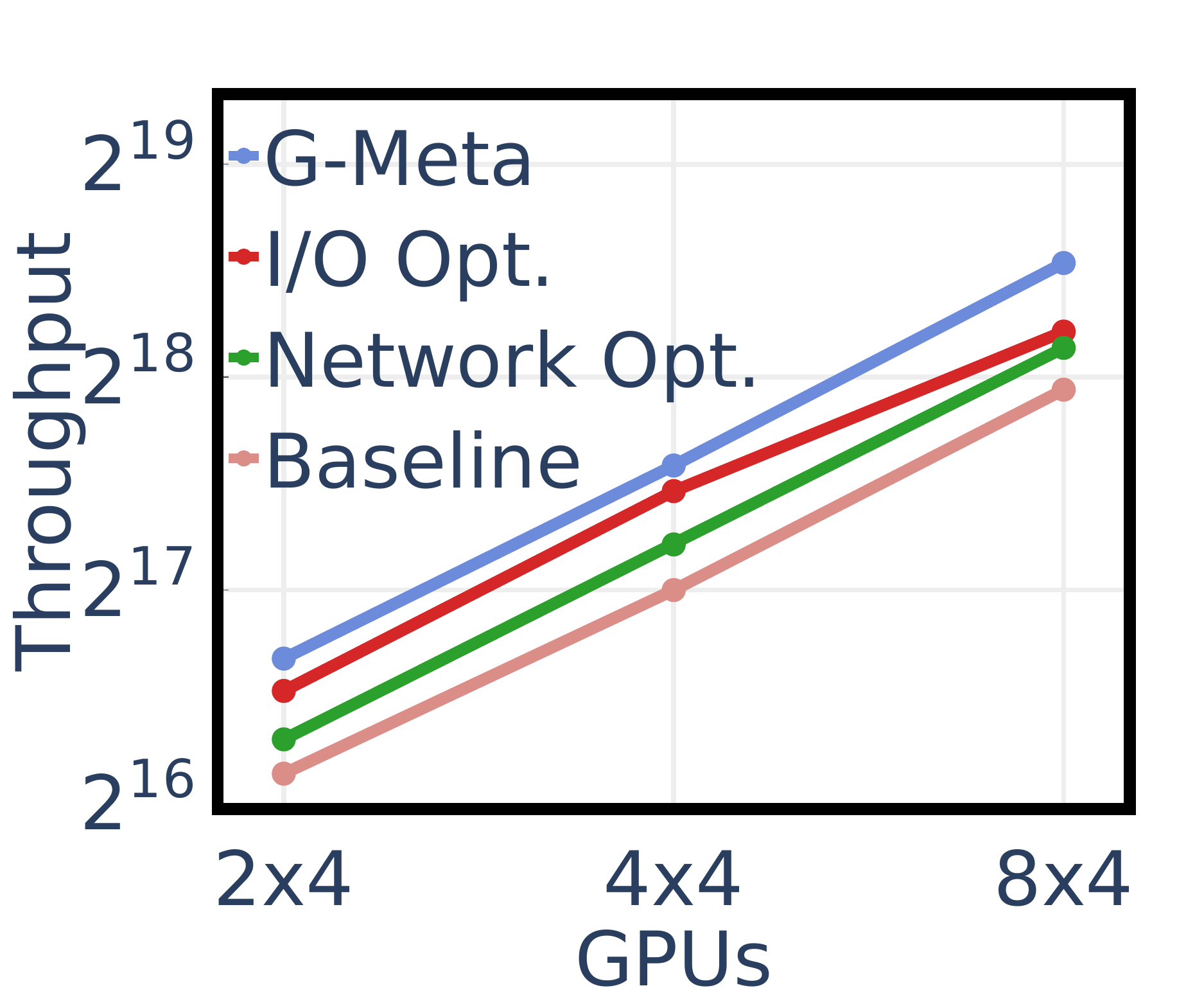}
\caption{The throughput given different experiment settings, like using I/O optimization or Network optimization in in-house data. }
\label{fig:ablation}
\end{minipage}%
\end{figure}

We perform the ablation study of both optimizations on I/O and network to further understand how they contribute to acceleration. Shown in Figure \ref{fig:ablation}, both optimizations on the network and I/O bring up 45\% and 51\% speedup compared with the baseline (G-Meta without these two optimizations) on $2\times4$ or $8\times4$ GPUs respectively. Specifically, I/O and network optimizations contribute to 27\% and 12 \% speedup respectively on $2\times4$ GPUs. However, the acceleration from I/O optimization shrinks on $8\times4$ GPUs. This is because the synchronous training has poor scalability with more participating nodes and the I/O stage in one node may block the whole iteration with high probability. Additionally, it is noteworthy that the baseline processes 72k samples per second on $2\times4$ GPUs, which approximates the same throughput (79k) using 80 workers  as displayed in Table ~\ref{tab:maml}. It shows that G-Meta with our algorithm is still efficient even without extra I/O and network optimizations.

\begin{table}[]
\caption{Average throughput (Samples/Seconds) and Speedup Ratio of G-Meta and DMAML using Ali-CCP and in-house dataset. Particularly, $2\times4$ means that there are 2 nodes and each node is equipped with 4 GPUs.} 
\label{tab:maml}
\begin{tabular}{llllll}
\hline
\textbf{}          & \multicolumn{4}{c}{\textbf{Throughput\textcolor{blue}{/Speedup Ratio}}} \\ \hline
\textbf{CPU workers}  & \textbf{20}           & \textbf{40}           & \textbf{80}           & \textbf{160}         \\ \hline
PS (public)        & 29k\textcolor{blue}{/1}     & 51k\textcolor{blue}{/0.88}      & 91k\textcolor{blue}{/0.78}     & 138k\textcolor{blue}{/0.59}     \\
PS (in-house)      & 27k\textcolor{blue}{/1}   & 48k\textcolor{blue}{/0.88}      & 79k\textcolor{blue}{/0.73}      & 126k\textcolor{blue}{/0.58}     \\ \hline
\textbf{GPU workers} & \textbf{\bm{$1 \times 4$}} & \textbf{\bm{$2 \times 4$}} & \textbf{\bm{$4 \times 4$}} & \textbf{\bm{$8 \times 4$}} \\ \hline
G-Meta (public)   & 90k\textcolor{blue}{/1}     & 169k\textcolor{blue}{/0.94}     & 322k\textcolor{blue}{/0.89}     & 618k\textcolor{blue}{/0.86}     \\
G-Meta (in-house) & 54k\textcolor{blue}{/1}     & 105k\textcolor{blue}{/0.97}    & 197k\textcolor{blue}{/0.91}     & 380k\textcolor{blue}{/0.88}     \\ \hline
\end{tabular}
\end{table}

\subsection{Online Deployment}
Since early 2022, G-Meta is widely used in Alipay's core applications. Using G-Meta remarkably shortens the overall training time by four times on average and ensures immediate delivery. Specifically, in our homepage display advertising, online experimental results show that the model delivery decreases from 3.7 to 1.2 hours using 1.6 billion records. The Conversion Rate (CVR) and Cost Per Mile (CPM) of the trained model respectively rise by 6.48\% and 1.06\%, benefiting from substantially more training data and tasks.

\section{Conclusion}
In this paper, we propose G-Meta for high-performance distributed training of optimization-based Meta DLRM models on GPU clusters. We design a hybrid parallelism algorithm with several optimizations to allow efficient distributed training of industrial Meta DLRM models over the GPU cluster. Our extensive experimental results and online deployment in several core applications clearly echo the efficiency and effectiveness of the method.

\bibliographystyle{ACM-Reference-Format}
\balance
\bibliography{distmaml.bib}


\begin{thebibliography}{39}


\ifx \showCODEN    \undefined \def \showCODEN     #1{\unskip}     \fi
\ifx \showDOI      \undefined \def \showDOI       #1{#1}\fi
\ifx \showISBNx    \undefined \def \showISBNx     #1{\unskip}     \fi
\ifx \showISBNxiii \undefined \def \showISBNxiii  #1{\unskip}     \fi
\ifx \showISSN     \undefined \def \showISSN      #1{\unskip}     \fi
\ifx \showLCCN     \undefined \def \showLCCN      #1{\unskip}     \fi
\ifx \shownote     \undefined \def \shownote      #1{#1}          \fi
\ifx \showarticletitle \undefined \def \showarticletitle #1{#1}   \fi
\ifx \showURL      \undefined \def \showURL       {\relax}        \fi
\providecommand\bibfield[2]{#2}
\providecommand\bibinfo[2]{#2}
\providecommand\natexlab[1]{#1}
\providecommand\showeprint[2][]{arXiv:#2}

\bibitem[Abadi et~al\mbox{.}(2016)]%
        {abadi2016tensorflow}
\bibfield{author}{\bibinfo{person}{Mart{\'\i}n Abadi}, \bibinfo{person}{Paul
  Barham}, \bibinfo{person}{Jianmin Chen}, \bibinfo{person}{Zhifeng Chen},
  \bibinfo{person}{Andy Davis}, \bibinfo{person}{Jeffrey Dean},
  \bibinfo{person}{Matthieu Devin}, \bibinfo{person}{Sanjay Ghemawat},
  \bibinfo{person}{Geoffrey Irving}, \bibinfo{person}{Michael Isard},
  {et~al\mbox{.}}} \bibinfo{year}{2016}\natexlab{}.
\newblock \showarticletitle{Tensorflow: a system for large-scale machine
  learning.}. In \bibinfo{booktitle}{\emph{Osdi}}, Vol.~\bibinfo{volume}{16}.
  Savannah, GA, USA, \bibinfo{pages}{265--283}.
\newblock


\bibitem[Aizman et~al\mbox{.}(2019)]%
        {aizman2019high}
\bibfield{author}{\bibinfo{person}{Alex Aizman}, \bibinfo{person}{Gavin
  Maltby}, {and} \bibinfo{person}{Thomas Breuel}.}
  \bibinfo{year}{2019}\natexlab{}.
\newblock \showarticletitle{High performance I/O for large scale deep
  learning}. In \bibinfo{booktitle}{\emph{2019 IEEE International Conference on
  Big Data (Big Data)}}. IEEE, \bibinfo{pages}{5965--5967}.
\newblock


\bibitem[Aliyun(2023)]%
        {aliyun}
\bibfield{author}{\bibinfo{person}{Aliyun}.} \bibinfo{year}{2023}\natexlab{}.
\newblock \bibinfo{booktitle}{\emph{Aliyun pricing}}.
\newblock
\urldef\tempurl%
\url{https://www.alibabacloud.com/pricing}
\showURL{%
Retrieved May 14, 2023 from \tempurl}


\bibitem[Bertinetto et~al\mbox{.}(2018)]%
        {bertinetto2018meta}
\bibfield{author}{\bibinfo{person}{Luca Bertinetto}, \bibinfo{person}{Joao~F
  Henriques}, \bibinfo{person}{Philip~HS Torr}, {and} \bibinfo{person}{Andrea
  Vedaldi}.} \bibinfo{year}{2018}\natexlab{}.
\newblock \showarticletitle{Meta-learning with differentiable closed-form
  solvers}.
\newblock \bibinfo{journal}{\emph{arXiv preprint arXiv:1805.08136}}
  (\bibinfo{year}{2018}).
\newblock


\bibitem[Bollenbacher et~al\mbox{.}(2020)]%
        {bollenbacher2020investigating}
\bibfield{author}{\bibinfo{person}{Jan Bollenbacher}, \bibinfo{person}{Florian
  Soulier}, \bibinfo{person}{Beate Rhein}, {and} \bibinfo{person}{Laurenz
  Wiskott}.} \bibinfo{year}{2020}\natexlab{}.
\newblock \showarticletitle{Investigating Parallelization of MAML}. In
  \bibinfo{booktitle}{\emph{International Conference on Discovery Science}}.
  Springer, \bibinfo{pages}{294--306}.
\newblock


\bibitem[Borthakur(2007)]%
        {borthakur2007hadoop}
\bibfield{author}{\bibinfo{person}{Dhruba Borthakur}.}
  \bibinfo{year}{2007}\natexlab{}.
\newblock \showarticletitle{The hadoop distributed file system: Architecture
  and design}.
\newblock \bibinfo{journal}{\emph{Hadoop Project Website}}
  \bibinfo{volume}{11}, \bibinfo{number}{2007} (\bibinfo{year}{2007}),
  \bibinfo{pages}{21}.
\newblock


\bibitem[Chen et~al\mbox{.}(2015)]%
        {chen2015mxnet}
\bibfield{author}{\bibinfo{person}{Tianqi Chen}, \bibinfo{person}{Mu Li},
  \bibinfo{person}{Yutian Li}, \bibinfo{person}{Min Lin},
  \bibinfo{person}{Naiyan Wang}, \bibinfo{person}{Minjie Wang},
  \bibinfo{person}{Tianjun Xiao}, \bibinfo{person}{Bing Xu},
  \bibinfo{person}{Chiyuan Zhang}, {and} \bibinfo{person}{Zheng Zhang}.}
  \bibinfo{year}{2015}\natexlab{}.
\newblock \showarticletitle{Mxnet: A flexible and efficient machine learning
  library for heterogeneous distributed systems}.
\newblock \bibinfo{journal}{\emph{arXiv preprint arXiv:1512.01274}}
  (\bibinfo{year}{2015}).
\newblock


\bibitem[Cheng et~al\mbox{.}(2016)]%
        {cheng2016wide}
\bibfield{author}{\bibinfo{person}{Heng-Tze Cheng}, \bibinfo{person}{Levent
  Koc}, \bibinfo{person}{Jeremiah Harmsen}, \bibinfo{person}{Tal Shaked},
  \bibinfo{person}{Tushar Chandra}, \bibinfo{person}{Hrishi Aradhye},
  \bibinfo{person}{Glen Anderson}, \bibinfo{person}{Greg Corrado},
  \bibinfo{person}{Wei Chai}, \bibinfo{person}{Mustafa Ispir}, {et~al\mbox{.}}}
  \bibinfo{year}{2016}\natexlab{}.
\newblock \showarticletitle{Wide \& deep learning for recommender systems}. In
  \bibinfo{booktitle}{\emph{Proceedings of the 1st workshop on deep learning
  for recommender systems}}. \bibinfo{pages}{7--10}.
\newblock


\bibitem[Dean et~al\mbox{.}(2012)]%
        {dean2012large}
\bibfield{author}{\bibinfo{person}{Jeffrey Dean}, \bibinfo{person}{Greg
  Corrado}, \bibinfo{person}{Rajat Monga}, \bibinfo{person}{Kai Chen},
  \bibinfo{person}{Matthieu Devin}, \bibinfo{person}{Mark Mao},
  \bibinfo{person}{Marc'aurelio Ranzato}, \bibinfo{person}{Andrew Senior},
  \bibinfo{person}{Paul Tucker}, \bibinfo{person}{Ke Yang}, {et~al\mbox{.}}}
  \bibinfo{year}{2012}\natexlab{}.
\newblock \showarticletitle{Large scale distributed deep networks}.
\newblock \bibinfo{journal}{\emph{Advances in neural information processing
  systems}}  \bibinfo{volume}{25} (\bibinfo{year}{2012}).
\newblock


\bibitem[Dean and Ghemawat(2008)]%
        {dean2008mapreduce}
\bibfield{author}{\bibinfo{person}{Jeffrey Dean} {and} \bibinfo{person}{Sanjay
  Ghemawat}.} \bibinfo{year}{2008}\natexlab{}.
\newblock \showarticletitle{MapReduce: simplified data processing on large
  clusters}.
\newblock \bibinfo{journal}{\emph{Commun. ACM}} \bibinfo{volume}{51},
  \bibinfo{number}{1} (\bibinfo{year}{2008}), \bibinfo{pages}{107--113}.
\newblock


\bibitem[Du et~al\mbox{.}(2019)]%
        {du2019sequential}
\bibfield{author}{\bibinfo{person}{Zhengxiao Du}, \bibinfo{person}{Xiaowei
  Wang}, \bibinfo{person}{Hongxia Yang}, \bibinfo{person}{Jingren Zhou}, {and}
  \bibinfo{person}{Jie Tang}.} \bibinfo{year}{2019}\natexlab{}.
\newblock \showarticletitle{Sequential scenario-specific meta learner for
  online recommendation}. In \bibinfo{booktitle}{\emph{Proceedings of the 25th
  ACM SIGKDD International Conference on Knowledge Discovery \& Data Mining}}.
  \bibinfo{pages}{2895--2904}.
\newblock


\bibitem[Finn et~al\mbox{.}(2017)]%
        {finn2017model}
\bibfield{author}{\bibinfo{person}{Chelsea Finn}, \bibinfo{person}{Pieter
  Abbeel}, {and} \bibinfo{person}{Sergey Levine}.}
  \bibinfo{year}{2017}\natexlab{}.
\newblock \showarticletitle{Model-agnostic meta-learning for fast adaptation of
  deep networks}. In \bibinfo{booktitle}{\emph{International conference on
  machine learning}}. PMLR, \bibinfo{pages}{1126--1135}.
\newblock


\bibitem[Gibiansky(2017)]%
        {allreduce}
\bibfield{author}{\bibinfo{person}{Andrew Gibiansky}.}
  \bibinfo{year}{2017}\natexlab{}.
\newblock \bibinfo{booktitle}{\emph{Bringing HPC Techniques to Deep Learning}}.
\newblock
\urldef\tempurl%
\url{https://andrew.gibiansky.com/blog/machine-learning/baidu-allreduce}
\showURL{%
Retrieved February 19, 2023 from \tempurl}


\bibitem[Harper and Konstan(2015)]%
        {harper2015movielens}
\bibfield{author}{\bibinfo{person}{F~Maxwell Harper} {and}
  \bibinfo{person}{Joseph~A Konstan}.} \bibinfo{year}{2015}\natexlab{}.
\newblock \showarticletitle{The movielens datasets: History and context}.
\newblock \bibinfo{journal}{\emph{Acm transactions on interactive intelligent
  systems (tiis)}} \bibinfo{volume}{5}, \bibinfo{number}{4}
  (\bibinfo{year}{2015}), \bibinfo{pages}{1--19}.
\newblock


\bibitem[Hospedales et~al\mbox{.}(2021)]%
        {hospedales2021meta}
\bibfield{author}{\bibinfo{person}{Timothy Hospedales},
  \bibinfo{person}{Antreas Antoniou}, \bibinfo{person}{Paul Micaelli}, {and}
  \bibinfo{person}{Amos Storkey}.} \bibinfo{year}{2021}\natexlab{}.
\newblock \showarticletitle{Meta-learning in neural networks: A survey}.
\newblock \bibinfo{journal}{\emph{IEEE transactions on pattern analysis and
  machine intelligence}} \bibinfo{volume}{44}, \bibinfo{number}{9}
  (\bibinfo{year}{2021}), \bibinfo{pages}{5149--5169}.
\newblock


\bibitem[Huan et~al\mbox{.}(2022)]%
        {huan2022industrial}
\bibfield{author}{\bibinfo{person}{Zhaoxin Huan}, \bibinfo{person}{Gongduo
  Zhang}, \bibinfo{person}{Xiaolu Zhang}, \bibinfo{person}{Jun Zhou},
  \bibinfo{person}{Qintong Wu}, \bibinfo{person}{Lihong Gu},
  \bibinfo{person}{Jinjie Gu}, \bibinfo{person}{Yong He}, \bibinfo{person}{Yue
  Zhu}, {and} \bibinfo{person}{Linjian Mo}.} \bibinfo{year}{2022}\natexlab{}.
\newblock \showarticletitle{An Industrial Framework for Cold-Start
  Recommendation in Zero-Shot Scenarios}. In
  \bibinfo{booktitle}{\emph{Proceedings of the 45th International ACM SIGIR
  Conference on Research and Development in Information Retrieval}}.
  \bibinfo{pages}{3403--3407}.
\newblock


\bibitem[Jeaugey(2017)]%
        {jeaugey2017nccl}
\bibfield{author}{\bibinfo{person}{Sylvain Jeaugey}.}
  \bibinfo{year}{2017}\natexlab{}.
\newblock \showarticletitle{Nccl 2.0}. In \bibinfo{booktitle}{\emph{GPU
  Technology Conference (GTC)}}, Vol.~\bibinfo{volume}{2}.
\newblock


\bibitem[Kunde et~al\mbox{.}(2021)]%
        {kunde2021distributed}
\bibfield{author}{\bibinfo{person}{Shruti Kunde}, \bibinfo{person}{Amey
  Pandit}, \bibinfo{person}{Mayank Mishra}, {and} \bibinfo{person}{Rekha
  Singhal}.} \bibinfo{year}{2021}\natexlab{}.
\newblock \showarticletitle{Distributed training for accelerating metalearning
  algorithms}. In \bibinfo{booktitle}{\emph{Proceedings of the International
  Workshop on Big Data in Emergent Distributed Environments}}.
  \bibinfo{pages}{1--6}.
\newblock


\bibitem[Lee et~al\mbox{.}(2019)]%
        {lee2019melu}
\bibfield{author}{\bibinfo{person}{Hoyeop Lee}, \bibinfo{person}{Jinbae Im},
  \bibinfo{person}{Seongwon Jang}, \bibinfo{person}{Hyunsouk Cho}, {and}
  \bibinfo{person}{Sehee Chung}.} \bibinfo{year}{2019}\natexlab{}.
\newblock \showarticletitle{Melu: Meta-learned user preference estimator for
  cold-start recommendation}. In \bibinfo{booktitle}{\emph{Proceedings of the
  25th ACM SIGKDD International Conference on Knowledge Discovery \& Data
  Mining}}. \bibinfo{pages}{1073--1082}.
\newblock


\bibitem[Li et~al\mbox{.}(2014)]%
        {li2014scaling}
\bibfield{author}{\bibinfo{person}{Mu Li}, \bibinfo{person}{David~G Andersen},
  \bibinfo{person}{Jun~Woo Park}, \bibinfo{person}{Alexander~J Smola},
  \bibinfo{person}{Amr Ahmed}, \bibinfo{person}{Vanja Josifovski},
  \bibinfo{person}{James Long}, \bibinfo{person}{Eugene~J Shekita}, {and}
  \bibinfo{person}{Bor-Yiing Su}.} \bibinfo{year}{2014}\natexlab{}.
\newblock \showarticletitle{Scaling distributed machine learning with the
  parameter server}. In \bibinfo{booktitle}{\emph{11th $\{$USENIX$\}$ Symposium
  on Operating Systems Design and Implementation ($\{$OSDI$\}$ 14)}}.
  \bibinfo{pages}{583--598}.
\newblock


\bibitem[Lu et~al\mbox{.}(2020)]%
        {lu2020meta}
\bibfield{author}{\bibinfo{person}{Yuanfu Lu}, \bibinfo{person}{Yuan Fang},
  {and} \bibinfo{person}{Chuan Shi}.} \bibinfo{year}{2020}\natexlab{}.
\newblock \showarticletitle{Meta-learning on heterogeneous information networks
  for cold-start recommendation}. In \bibinfo{booktitle}{\emph{Proceedings of
  the 26th ACM SIGKDD International Conference on Knowledge Discovery \& Data
  Mining}}. \bibinfo{pages}{1563--1573}.
\newblock


\bibitem[Ma et~al\mbox{.}(2018)]%
        {ma2018entire}
\bibfield{author}{\bibinfo{person}{Xiao Ma}, \bibinfo{person}{Liqin Zhao},
  \bibinfo{person}{Guan Huang}, \bibinfo{person}{Zhi Wang},
  \bibinfo{person}{Zelin Hu}, \bibinfo{person}{Xiaoqiang Zhu}, {and}
  \bibinfo{person}{Kun Gai}.} \bibinfo{year}{2018}\natexlab{}.
\newblock \showarticletitle{Entire space multi-task model: An effective
  approach for estimating post-click conversion rate}. In
  \bibinfo{booktitle}{\emph{The 41st International ACM SIGIR Conference on
  Research \& Development in Information Retrieval}}.
  \bibinfo{pages}{1137--1140}.
\newblock


\bibitem[Ma et~al\mbox{.}(2020)]%
        {ma2020temporal}
\bibfield{author}{\bibinfo{person}{Yifei Ma}, \bibinfo{person}{Balakrishnan
  Narayanaswamy}, \bibinfo{person}{Haibin Lin}, {and} \bibinfo{person}{Hao
  Ding}.} \bibinfo{year}{2020}\natexlab{}.
\newblock \showarticletitle{Temporal-contextual recommendation in real-time}.
  In \bibinfo{booktitle}{\emph{Proceedings of the 26th ACM SIGKDD international
  conference on knowledge discovery \& data mining}}.
  \bibinfo{pages}{2291--2299}.
\newblock


\bibitem[Mudigere et~al\mbox{.}(2022)]%
        {mudigere2022software}
\bibfield{author}{\bibinfo{person}{Dheevatsa Mudigere}, \bibinfo{person}{Yuchen
  Hao}, \bibinfo{person}{Jianyu Huang}, \bibinfo{person}{Zhihao Jia},
  \bibinfo{person}{Andrew Tulloch}, \bibinfo{person}{Srinivas Sridharan},
  \bibinfo{person}{Xing Liu}, \bibinfo{person}{Mustafa Ozdal},
  \bibinfo{person}{Jade Nie}, \bibinfo{person}{Jongsoo Park}, {et~al\mbox{.}}}
  \bibinfo{year}{2022}\natexlab{}.
\newblock \showarticletitle{Software-hardware co-design for fast and scalable
  training of deep learning recommendation models}. In
  \bibinfo{booktitle}{\emph{Proceedings of the 49th Annual International
  Symposium on Computer Architecture}}. \bibinfo{pages}{993--1011}.
\newblock


\bibitem[Nichol et~al\mbox{.}(2018)]%
        {nichol2018first}
\bibfield{author}{\bibinfo{person}{Alex Nichol}, \bibinfo{person}{Joshua
  Achiam}, {and} \bibinfo{person}{John Schulman}.}
  \bibinfo{year}{2018}\natexlab{}.
\newblock \showarticletitle{On first-order meta-learning algorithms}.
\newblock \bibinfo{journal}{\emph{arXiv preprint arXiv:1803.02999}}
  (\bibinfo{year}{2018}).
\newblock


\bibitem[Pan et~al\mbox{.}(2019)]%
        {pan2019warm}
\bibfield{author}{\bibinfo{person}{Feiyang Pan}, \bibinfo{person}{Shuokai Li},
  \bibinfo{person}{Xiang Ao}, \bibinfo{person}{Pingzhong Tang}, {and}
  \bibinfo{person}{Qing He}.} \bibinfo{year}{2019}\natexlab{}.
\newblock \showarticletitle{Warm up cold-start advertisements: Improving ctr
  predictions via learning to learn id embeddings}. In
  \bibinfo{booktitle}{\emph{Proceedings of the 42nd International ACM SIGIR
  Conference on Research and Development in Information Retrieval}}.
  \bibinfo{pages}{695--704}.
\newblock


\bibitem[Paszke et~al\mbox{.}(2019)]%
        {paszke2019pytorch}
\bibfield{author}{\bibinfo{person}{Adam Paszke}, \bibinfo{person}{Sam Gross},
  \bibinfo{person}{Francisco Massa}, \bibinfo{person}{Adam Lerer},
  \bibinfo{person}{James Bradbury}, \bibinfo{person}{Gregory Chanan},
  \bibinfo{person}{Trevor Killeen}, \bibinfo{person}{Zeming Lin},
  \bibinfo{person}{Natalia Gimelshein}, \bibinfo{person}{Luca Antiga},
  {et~al\mbox{.}}} \bibinfo{year}{2019}\natexlab{}.
\newblock \showarticletitle{Pytorch: An imperative style, high-performance deep
  learning library}.
\newblock \bibinfo{journal}{\emph{Advances in neural information processing
  systems}}  \bibinfo{volume}{32} (\bibinfo{year}{2019}).
\newblock


\bibitem[Rajeswaran et~al\mbox{.}(2019)]%
        {rajeswaran2019meta}
\bibfield{author}{\bibinfo{person}{Aravind Rajeswaran},
  \bibinfo{person}{Chelsea Finn}, \bibinfo{person}{Sham~M Kakade}, {and}
  \bibinfo{person}{Sergey Levine}.} \bibinfo{year}{2019}\natexlab{}.
\newblock \showarticletitle{Meta-learning with implicit gradients}.
\newblock \bibinfo{journal}{\emph{Advances in neural information processing
  systems}}  \bibinfo{volume}{32} (\bibinfo{year}{2019}).
\newblock


\bibitem[Ravi and Larochelle(2017)]%
        {ravi2017optimization}
\bibfield{author}{\bibinfo{person}{Sachin Ravi} {and} \bibinfo{person}{Hugo
  Larochelle}.} \bibinfo{year}{2017}\natexlab{}.
\newblock \showarticletitle{Optimization as a model for few-shot learning}. In
  \bibinfo{booktitle}{\emph{International conference on learning
  representations}}.
\newblock


\bibitem[Sergeev and Del~Balso(2018)]%
        {sergeev2018horovod}
\bibfield{author}{\bibinfo{person}{Alexander Sergeev} {and}
  \bibinfo{person}{Mike Del~Balso}.} \bibinfo{year}{2018}\natexlab{}.
\newblock \showarticletitle{Horovod: fast and easy distributed deep learning in
  TensorFlow}.
\newblock \bibinfo{journal}{\emph{arXiv preprint arXiv:1802.05799}}
  (\bibinfo{year}{2018}).
\newblock


\bibitem[Song et~al\mbox{.}(2021)]%
        {song2021cbml}
\bibfield{author}{\bibinfo{person}{Jiayu Song}, \bibinfo{person}{Jiajie Xu},
  \bibinfo{person}{Rui Zhou}, \bibinfo{person}{Lu Chen},
  \bibinfo{person}{Jianxin Li}, {and} \bibinfo{person}{Chengfei Liu}.}
  \bibinfo{year}{2021}\natexlab{}.
\newblock \showarticletitle{CBML: A cluster-based meta-learning model for
  session-based recommendation}. In \bibinfo{booktitle}{\emph{Proceedings of
  the 30th ACM International Conference on Information \& Knowledge
  Management}}. \bibinfo{pages}{1713--1722}.
\newblock


\bibitem[Tensorflow(2023)]%
        {tfrecords}
\bibfield{author}{\bibinfo{person}{Tensorflow}.}
  \bibinfo{year}{2023}\natexlab{}.
\newblock \bibinfo{booktitle}{\emph{TFRecords}}.
\newblock
\urldef\tempurl%
\url{https://www.tensorflow.org/tutorials/load_data/tfrecord}
\showURL{%
Retrieved February 19, 2023 from \tempurl}


\bibitem[Vartak et~al\mbox{.}(2017)]%
        {vartak2017meta}
\bibfield{author}{\bibinfo{person}{Manasi Vartak}, \bibinfo{person}{Arvind
  Thiagarajan}, \bibinfo{person}{Conrado Miranda}, \bibinfo{person}{Jeshua
  Bratman}, {and} \bibinfo{person}{Hugo Larochelle}.}
  \bibinfo{year}{2017}\natexlab{}.
\newblock \showarticletitle{A meta-learning perspective on cold-start
  recommendations for items}.
\newblock \bibinfo{journal}{\emph{Advances in neural information processing
  systems}}  \bibinfo{volume}{30} (\bibinfo{year}{2017}).
\newblock


\bibitem[Volkovs et~al\mbox{.}(2017)]%
        {volkovs2017dropoutnet}
\bibfield{author}{\bibinfo{person}{Maksims Volkovs}, \bibinfo{person}{Guangwei
  Yu}, {and} \bibinfo{person}{Tomi Poutanen}.} \bibinfo{year}{2017}\natexlab{}.
\newblock \showarticletitle{Dropoutnet: Addressing cold start in recommender
  systems}.
\newblock \bibinfo{journal}{\emph{Advances in neural information processing
  systems}}  \bibinfo{volume}{30} (\bibinfo{year}{2017}).
\newblock


\bibitem[Vuorio et~al\mbox{.}(2019)]%
        {vuorio2019multimodal}
\bibfield{author}{\bibinfo{person}{Risto Vuorio}, \bibinfo{person}{Shao-Hua
  Sun}, \bibinfo{person}{Hexiang Hu}, {and} \bibinfo{person}{Joseph~J Lim}.}
  \bibinfo{year}{2019}\natexlab{}.
\newblock \showarticletitle{Multimodal model-agnostic meta-learning via
  task-aware modulation}.
\newblock \bibinfo{journal}{\emph{Advances in neural information processing
  systems}}  \bibinfo{volume}{32} (\bibinfo{year}{2019}).
\newblock


\bibitem[Wang et~al\mbox{.}(2022)]%
        {wang2022merlin}
\bibfield{author}{\bibinfo{person}{Zehuan Wang}, \bibinfo{person}{Yingcan Wei},
  \bibinfo{person}{Minseok Lee}, \bibinfo{person}{Matthias Langer},
  \bibinfo{person}{Fan Yu}, \bibinfo{person}{Jie Liu}, \bibinfo{person}{Shijie
  Liu}, \bibinfo{person}{Daniel~G Abel}, \bibinfo{person}{Xu Guo},
  \bibinfo{person}{Jianbing Dong}, {et~al\mbox{.}}}
  \bibinfo{year}{2022}\natexlab{}.
\newblock \showarticletitle{Merlin HugeCTR: GPU-accelerated Recommender System
  Training and Inference}. In \bibinfo{booktitle}{\emph{Proceedings of the 16th
  ACM Conference on Recommender Systems}}. \bibinfo{pages}{534--537}.
\newblock


\bibitem[Yang et~al\mbox{.}(2022)]%
        {yang2022task}
\bibfield{author}{\bibinfo{person}{Jieyu Yang}, \bibinfo{person}{Zhaoxin Huan},
  \bibinfo{person}{Yong He}, \bibinfo{person}{Ke Ding}, \bibinfo{person}{Liang
  Zhang}, \bibinfo{person}{Xiaolu Zhang}, \bibinfo{person}{Jun Zhou}, {and}
  \bibinfo{person}{Linjian Mo}.} \bibinfo{year}{2022}\natexlab{}.
\newblock \showarticletitle{Task Similarity Aware Meta Learning for Cold-Start
  Recommendation}. In \bibinfo{booktitle}{\emph{Proceedings of the 31st ACM
  International Conference on Information \& Knowledge Management}}.
  \bibinfo{pages}{4630--4634}.
\newblock


\bibitem[Zhang et~al\mbox{.}(2022)]%
        {zhang2022picasso}
\bibfield{author}{\bibinfo{person}{Yuanxing Zhang}, \bibinfo{person}{Langshi
  Chen}, \bibinfo{person}{Siran Yang}, \bibinfo{person}{Man Yuan},
  \bibinfo{person}{Huimin Yi}, \bibinfo{person}{Jie Zhang},
  \bibinfo{person}{Jiamang Wang}, \bibinfo{person}{Jianbo Dong},
  \bibinfo{person}{Yunlong Xu}, \bibinfo{person}{Yue Song}, {et~al\mbox{.}}}
  \bibinfo{year}{2022}\natexlab{}.
\newblock \showarticletitle{PICASSO: Unleashing the Potential of GPU-centric
  Training for Wide-and-deep Recommender Systems}.
\newblock \bibinfo{journal}{\emph{arXiv preprint arXiv:2204.04903}}
  (\bibinfo{year}{2022}).
\newblock


\bibitem[Zhou et~al\mbox{.}(2019)]%
        {zhou2019deep}
\bibfield{author}{\bibinfo{person}{Guorui Zhou}, \bibinfo{person}{Na Mou},
  \bibinfo{person}{Ying Fan}, \bibinfo{person}{Qi Pi}, \bibinfo{person}{Weijie
  Bian}, \bibinfo{person}{Chang Zhou}, \bibinfo{person}{Xiaoqiang Zhu}, {and}
  \bibinfo{person}{Kun Gai}.} \bibinfo{year}{2019}\natexlab{}.
\newblock \showarticletitle{Deep interest evolution network for click-through
  rate prediction}. In \bibinfo{booktitle}{\emph{Proceedings of the AAAI
  conference on artificial intelligence}}, Vol.~\bibinfo{volume}{33}.
  \bibinfo{pages}{5941--5948}.
\newblock


\end{thebibliography}

\end{document}